**Title:** Instant and Reversible Adhesive-free Bonding Between Silicones and Glossy Papers for Soft Robotics

**Authors:**

Takumi Shibuya[1], Kazuya Murakami[1], Akitsu Shigetou[2], and Jun Shintake[1]*

**Affiliations:**

[1]Department of Mechanical and Intelligent Systems Engineering, The University of Electro-communications, 1-5-1 Chofugaoka, Chofu, Tokyo 182-8585, Japan.

[2]Research Center for Materials Nanoarchitectonics, National Institute for Materials Science, Tsukuba, Ibaraki 305-0044, Japan.

*E-mail: shintake@uec.ac.jp

**Abstract:**

Integrating silicone with non-extensible materials is a common strategy used in the fabrication of fluidically-driven soft actuators, yet conventional approaches often rely on irreversible adhesives or embedding processes that are labor-intensive and difficult to modify. This work presents silicone–glossy paper bonding (SGB), a rapid, adhesive-free, and solvent-reversible bonding approach that forms robust silicone-paper interfaces simply through contact. The SGB interface withstands high mechanical loads (shear strength > 61 kPa) and can be fully detached and reassembled via ethanol immersion without loss of performance, enabling component reuse and rapid redesign. Characterization studies indicate that surface functional groups primarily govern adhesion on the glossy paper and the modulus of the silicone, while durability and environmental response clarify the conditions for reversible debonding. The results further suggest a synergistic interaction of hydrogen bonding and oligomer diffusion, yielding strong yet reconfigurable adhesion. Soft actuators fabricated using SGB design exhibit equal or greater performance compared to conventional embedded-layer design and enable programmable actuation modes, including contraction, bending, and twisting. By simplifying fabrication while supporting reuse and rapid iteration, SGB offers a scalable and sustainable platform for rapid prototyping in soft robotics.

**Introduction**

Actuators fabricated from compliant or flexible materials are essential for soft robots, enabling safe interaction with living organisms including humans, adaptation to surrounding objects and environments, and the expression of life-like locomotion.[1–4] These characteristics have led to diverse applications in industry,[5,6] healthcare,[7,8] disaster response[9–11], environmental monitoring,[12,13] and daily life.[14,15] Among the various soft actuator designs, fluidically-driven systems, whether pneumatic[16] or hydraulic[17], are particularly prevalent because of their structural simplicity and ease of implementation. Silicones are widely used in these actuators due to their high compliance, chemical stability, thermal resistance, ease of handling, and customizability.[10,18–20] A common fabrication method for fluidically-driven soft actuators is to combine silicone with non-extensible materials such as paper, fabric, or fibers.[21–26] By constraining specific regions of the silicone structure, this method enables versatile and complex motions such as bending, twisting, or combinations thereof, rather than simple isotropic expansion.[22–25]

However, integrating silicone with non-extensible materials presents significant manufacturing challenges. Conventional approaches include using specialized adhesives for substrates that are difficult to bond or embedding the materials into the silicone matrix before curing.[22,25,27] However, these approaches have several drawbacks. First, the use of adhesives and the embedding process complicate the fabrication workflow, as the curing times of the adhesive and raw silicone extend the overall production duration. Second, the cured adhesive introduces rigidity, potentially compromising the actuators' inherent flexibility and degrading their performance. Third, embedding non-extensible materials requires additional silicone matrix to accommodate them, leading to an increase in volume that restricts design freedom. Fourth, both adhesion and embedding are typically irreversible processes, hindering the reuse, reprocessing, or replacement of components. These challenges could also complicate the integration of external components such as sensors.

In this study, we present a method for integrating silicone with non-extensible materials to fabricate soft actuators that exploit an adhesion phenomenon occurring between silicone and glossy papers. This method achieves instant and reversible bonding upon contact, eliminating the need for adhesives and an additional silicone matrix. Consequently, it significantly simplifies the fabrication process, reduces production time, preserves the inherent flexibility of the actuators, minimizes restrictions on design freedom, and enables easy detachment and reconfiguration of components. In the following sections, we provide an overview of the method and investigate the bonding characteristics. Furthermore, we analyze the properties of glossy papers and the bonding interface to explore the underlying adhesion mechanism. Finally, we demonstrate how the proposed method can be applied to fluidically-driven soft actuators.

**Results**

**Silicone-Glossy Paper Bonding (SGB)**

We term the adhesion phenomenon as silicone-glossy paper bonding (SGB), characterized by instant and robust locking between specific silicones and glossy paper upon contact (Fig. 1a, see also Supplementary Video S1). Tensile testing revealed that the SGB interface exhibits adhesion strength exceeding the bulk strength of the silicone, as failure occurred within the silicone rather than at the interface. This strong bonding remained stable even under dynamic loading (Supplementary Video S2). The macroscopic reliability of SGB is demonstrated in a load-bearing test, in which a silicone-glossy paper composite loop with a bonding area of 89 mm × 90 mm supports a human weight of ~50 kg (Fig. 1b, see also Supplementary Video S3). These results highlight the load-bearing capability of SGB for scalable applications in soft robotics. Importantly, SGB offers chemically induced reversibility, enabling non-destructive reconfiguration of a soft pneumatic actuator using ethanol (Fig. 1c). The SGB interface features sealed chambers that can sustain pneumatic pressure for bending actuation (see Supplementary Video S4). Immersion in ethanol for 15 min enables clean debonding without damaging the substrate. After ethanol evaporation, the actuator can be successfully reassembled and operated again. Unlike hydrogels, SGB leverages the chemical stability of silicone to provide a durable, reversible, and reconfigurable bonding strategy for soft robotic architectures.

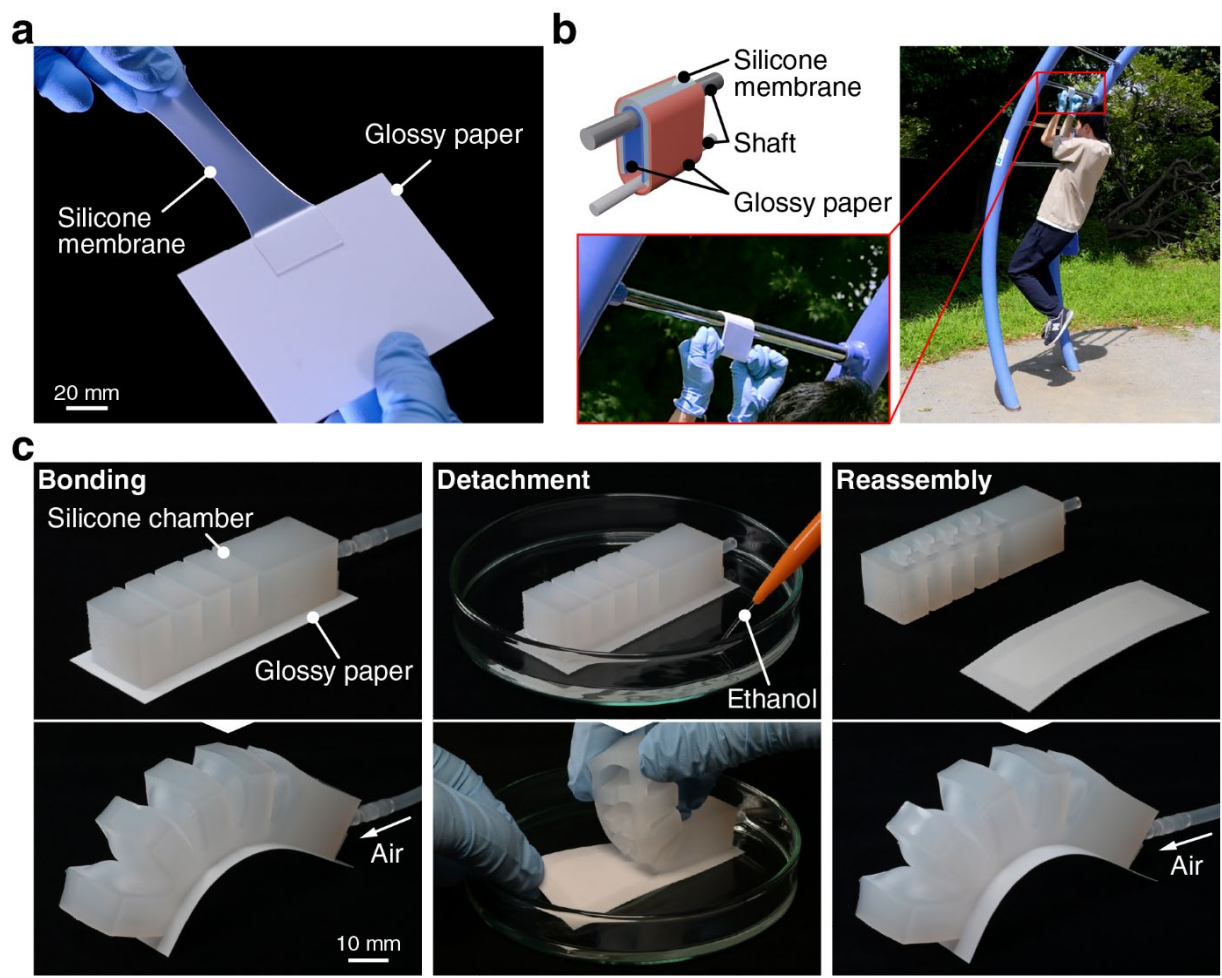

**Fig. 1: Overview of silicone-glossy paper bonding (SGB). a** Formation of the SGB interface, showing instant bonding between a cured silicone membrane and glossy paper without external adhesives or additional processing. **b** Load-bearing demonstration, where a silicone–glossy paper composite loop with a bonding area of 89 mm × 90 mm supports a human weight (~50 kg) without delamination. **c** Chemically induced reversibility applied to a soft pneumatic actuator: the SGB-bonded actuator generates bending motion under pneumatic pressure; immersion in ethanol enables clean detachment into individual components; and drying and reassembly restore full actuator functionality.

**Influence of Constituent Materials**

Understanding the reversible bonding characteristics of SGB is essential for evaluating its potential applications. To characterize the interfacial bond between silicone and glossy paper, we employ T-peel testing (Fig. 2a). The test specimen consists of a silicone membrane sandwiched between two pieces of glossy paper, allowing for the assessment of adhesive interaction under varying experimental conditions (Fig. S1, Supplementary Information).

To assess the influence of substrate composition, we evaluate adhesion strength across seven commercially available glossy papers (P1–P7$_b$) paired with Ecoflex 00-30 (Smooth-On) (Fig. 2b). P1 yields the highest adhesion strength of 2.8 N mm$^{-1}$, whereas P2–P6 cluster between 0.15 and 0.3 N mm$^{-1}$. Conversely, P7$_a$ and P7$_b$ display negligible adhesion (~0.05 N mm$^{-1}$), likely reflecting only the inherent tack of the silicone rather than specific interfacial bonding. These results underscore the substrate type as a critical design parameter. We further examine the role of silicone modulus on SGB strength by varying the softener content in Dragon Skin 10 Medium (Smooth-On) (Fig. 2c). Adhesion strength positively correlates with softener concentration, reaching 0.94 N mm$^{-1}$ at 0%, 2.5 N mm$^{-1}$ at 25%, and 2.9 N mm$^{-1}$ at 50%. Notably, at 50% softener addition, the modulus of Dragon Skin becomes comparable to that of Ecoflex 00-30, resulting in a nearly identical peak adhesion strength (2.8 N mm$^{-1}$). This suggests that silicone hardness is a key factor in determining adhesion performance. Consequently, we adopt the P1 substrate and Ecoflex 00-30 for subsequent characterization.

**Environmental Stability and Mechanical Durability**

To validate the switchable adhesion capability of SGB, we assess its reversibility and environmental adaptability. We first confirm the separation and re-adhesion process of the glossy paper substrate P1 and Ecoflex 00-30 using ethanol in a fluidically-driven soft actuator configuration (Fig. 1c). We then quantify the impact of immersion time on the adhesion strength in deionized water, ethanol, and tap water (Fig. 2d). Immersion for 1 min attenuated the adhesion strength from 2.8 to 1.4, 1.1, and 1.3 N mm$^{-1}$ for deionized water, ethanol, and tap water, respectively. This substantial reduction (>50% of the initial value) is advantageous, as it enables low-stress detachment of the interface. While ethanol and tap water induce an asymptotic decrease in strength over time with minimum values of 0.11 and 0.16, respectively, deionized water yielded non-monotonic fluctuations (1.2–2.0 N mm$^{-1}$), suggesting a distinct saturation mechanism. Importantly, demonstrating high reusability, the specimen re-bonded after ethanol-mediated separation, achieving an adhesion strength of 3.4 N mm$^{-1}$. This confirms that SGB exhibits a robust, reversible bonding mechanism, in which liquid-induced separation does not compromise the subsequent structural integrity (Fig. S2, Supplementary Information).

Thermal stability, a prerequisite for soft robotics,[10] is then evaluated by measuring adhesion at elevated temperatures (60, 120, and 200°C) (Fig. 2e). Adhesion strength decreases with increasing temperature, presumably due to thermal softening or degradation of the glossy paper's coating layer. Nevertheless, substantial bonding (0.87 N mm$^{-1}$) persists even at 200°C, demonstrating the applicability of SGB in high-temperature environments when appropriate safety margins are observed.

Finally, we characterize cyclic fatigue behavior over 200 displacement cycles to determine practical durability (Fig. 2f). The initial strength of 2.2 N mm$^{-1}$ (78% of maximum) drops sharply to 0.97 N mm$^{-1}$ (46%) within 10 cycles, followed by a gradual decline to 0.20 N mm$^{-1}$ (9.0%) at 200 cycles. Post-test analysis reveals predominantly cohesive failure within the silicone rather than interfacial delamination. This behavior suggests that the initial steep decline arises from ductile damage or fatigue-induced crack initiation within the silicone bulk, whereas the later plateau corresponds to slower crack propagation. Thus, the fatigue life of the silicone itself governs the long-term durability of SGB-based devices.

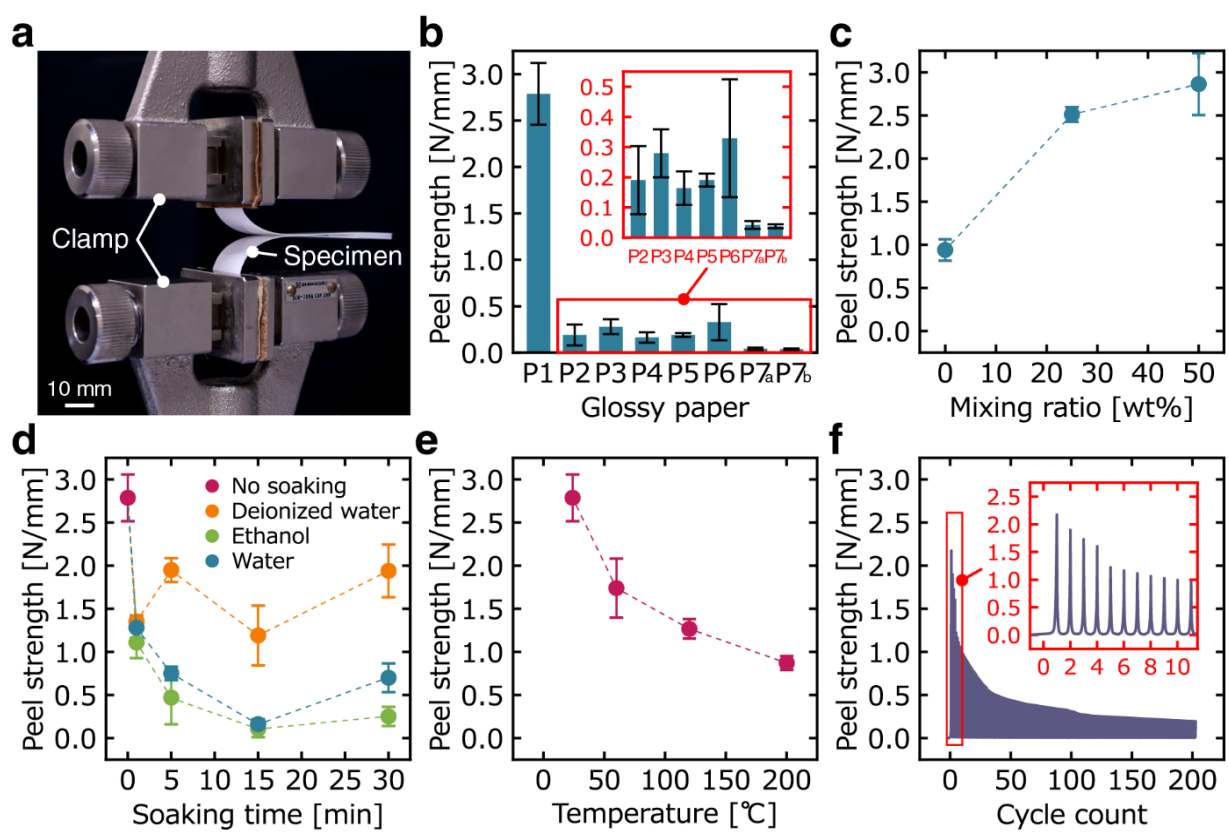

**Fig. 2: Adhesion characterization of silicone–glossy paper bonding (SGB). a** Schematic illustration of the peel-test setup. **b** Adhesion strength for various glossy papers. **c** Effect of plasticizer ratio in Dragon Skin 10 MEDIUM on adhesion performance. **d** Influence of chemical immersion on adhesion strength. **e** Temperature dependence of adhesion strength. **f** Durability under cyclic displacement. Values in **b**–**e** represent the mean and the standard deviation ($N = 3$).

**Adhesion Mechanism**

Understanding the underlying adhesion principles is crucial for predicting device performance and identifying the scope of potential applications. Silicone is generally chemically inert and exhibits low surface energy, characteristics that typically lead to poor adhesion.[28] As a result, bonding silicone usually requires surface activation processes, such as plasma treatment or the use of specialized primers and adhesives [29–33]. Conversely, SGB achieves instantaneous adhesion simply by bringing the silicone and glossy paper into contact. The absence of this phenomenon in conventional bonding strategies suggests that rapid chemical interactions or dominant physical processes are involved. To elucidate the adhesion mechanism, we synthesize the characterization results presented in the preceding sections and highlight three key findings that govern SGB behavior.

The first key finding is that adhesion strength varies widely across glossy papers, exhibiting nearly an order-of-magnitude variation across substrates. To investigate the source of this disparity, we examine surface roughness and chemical composition. Surface profilometry reveals that all papers possess similar roughness values ($R_a$ ~ 100–250 nm), showing no correlation with adhesion strength (Fig. 3a). X-ray Photoelectron Spectroscopy (XPS) detects O and C in all samples (Fig. 3b); Al appears in P4, while Si is detected in the others, likely originating from alumina and silica used in ink-receiving layers.[34] However, these elemental differences do not align with the adhesion trends (Fig. 2b), indicating that the bulk coating composition is not the primary driver. Consequently, we focus on surface functional groups. Narrow-scan C1s spectra exhibit peaks at 284.8 eV (C–C/C–H) and 288.8 eV (O=C–O), indicating the presence of carboxyl (–COOH) or ester (–COOR) groups.[35] P1—the paper exhibiting the strongest adhesion—displays a broader peak. Peak deconvolution confirms C–O bonds, indicating the presence of hydroxyl (–OH) groups or ether (C–O–C) linkages (Fig. 3c).[35] These –OH groups, which may have been introduced during surface treatments for ink fixation, likely contribute substantially to the superior adhesion exhibited by P1.[36] To further characterize the interface, SEM imaging confirms intimate contact without observable gaps (Fig. 3d). TEM combined with EELS mapping (50 × 10 points) of Si, C, and O captures the interfacial region (Fig. 3e and 3f). An EELS line profile extracted along the P–S trajectory reveals three notable features: diffusion of silicon species into the glossy paper layer, a chemical shift of the Si edge peak to approximately 108 eV, indicating an oxide-like bonding state (Si–O or Si=O),[37,38] and a C-K edge signature of the silicone side resembling "CHO"-type species akin to hydrated amorphous carbon (Fig. 3g). These results support a coupled chemical–physical adhesion mechanism.

The second key finding concerns how solvent immersion reversibly reduces adhesion. After 1 min of immersion, adhesion strength decreases to ~50% of its initial value in all solutions, likely due to water infiltration from specimen edges through capillary action. Beyond 5 min, distinct behaviors emerge: adhesion continues to decrease in ethanol and tap water, whereas the decrease plateaus in deionized water. This divergence likely reflects differences in interfacial polarity. In ethanol and tap water, hydrolysis is promoted at the crack tip, and the newly exposed surfaces strongly adsorb and reversibly bind the liquid, which stabilizes the infiltrated liquid and sustains crack formation. In the case of deionized water, the liquid lacks sufficient ionic or polar environment to disrupt the interfacial interactions fully, and its high oxidative potential oxidizes the silicone surface, suppresses reversible hydrolysis, and makes surface adsorption energy and capillary infiltration rate limiting. Combining these insights with the first key finding, we conclude that a synergy of chemical and physical interactions governs SGB. Chemically, functional groups on the silicone surface can form hydrogen bonds with components of the glossy paper, followed by dehydration polymerization that strengthens the bond. Although dehydration condensation between methyl groups is unlikely at room temperature, hydrogen bonding involving esterified carboxyl groups can readily occur. Physically, fragmented, amorphous-like silicone oligomers may diffuse into the paper microstructure, contributing to mechanical interlocking. Reversibility arises because solvent immersion induces hydrolysis, cleaving these bonds. Once the solvent evaporates, the surfaces return to an esterified state, enabling the bond to reform upon recontact.

The third key finding establishes that the mechanical properties of silicone significantly influence adhesion. Softer silicones increase the real contact area with the glossy paper surface, enhancing intermolecular forces such as hydrogen bonding and van der Waals interactions. Their flexibility also enables stress relaxation during peeling, reducing interfacial stress concentration.[39] As a result, SGB achieves a unique combination of high adhesion strength and reconfigurability, with silicone hardness serving as a critical design parameter.

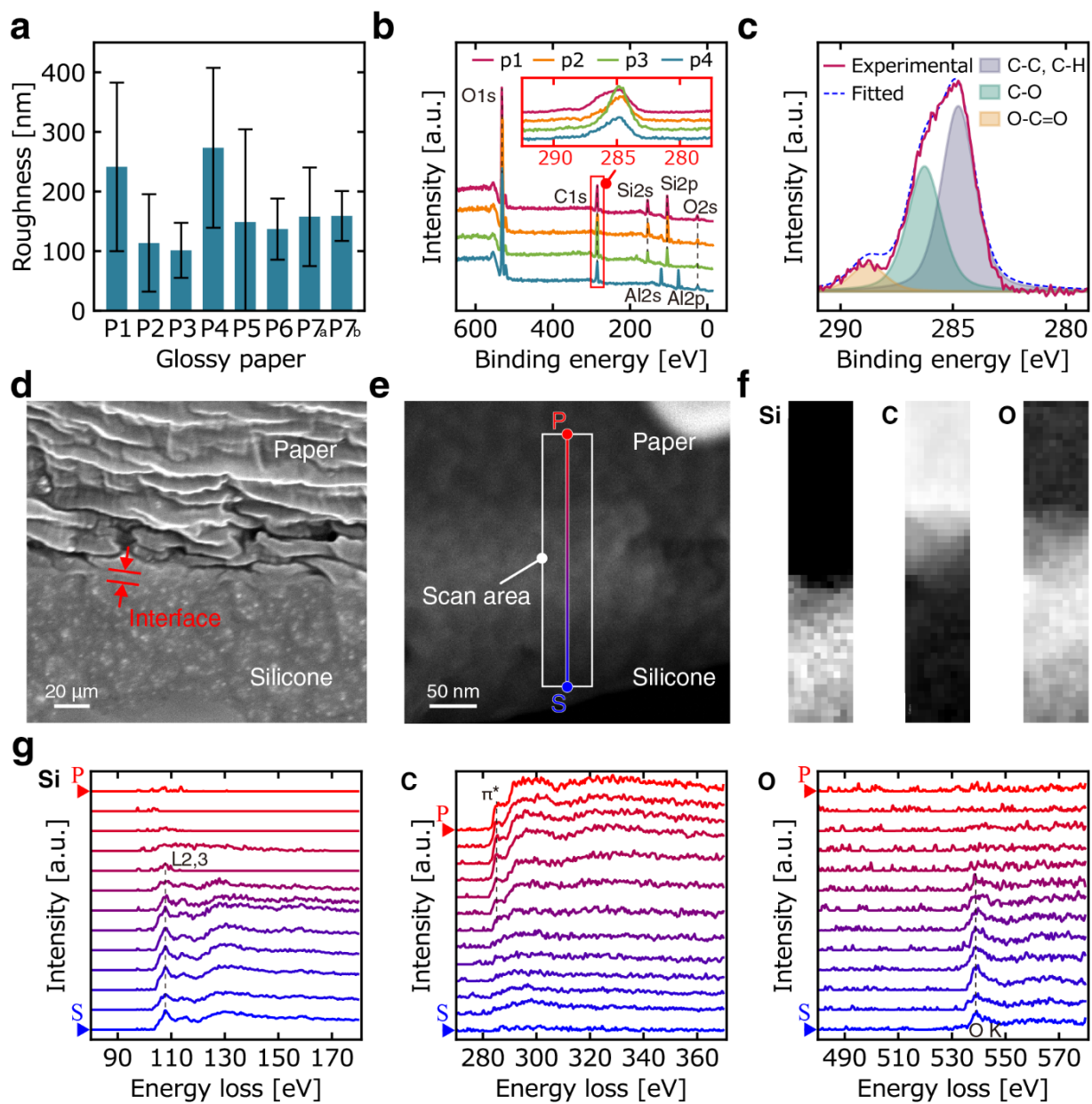

**Fig. 3: Surface and interface analysis of silicone–glossy paper bonding (SGB). a** Surface roughness of glossy papers. Values represent the mean and the standard deviation ($N = 6$). **b** XPS surface binding spectra of glossy papers. **c** Deconvoluted C1s peaks for P1. **d** SEM image of the adhesive interface. **e** TEM-STEM image and corresponding STEM–EELS scanning region. **f** EELS elemental mapping for Si, C, and O. **g** Elemental profiles for Si, C, and O across the interface.

**SGB-based Fluidically-Driven Soft Actuators**

To demonstrate the effectiveness of SGB for soft robotic devices, we first fabricate and characterize an SGB-based fluidically-driven soft actuator, comparing its performance with that of an actuator fabricated using a conventional method (Fig. 4a). In this method, the paper is embedded within the silicone layer (Fig. S3, Supplementary Information).[25] For the SGB-based actuator, both the tip angle and blocking force increase with rising input pressure (Fig. 4b and 4c), with an overall trend comparable to that of the conventional actuator. This confirms that SGB is compatible with devices undergoing large deformations. Although tip angle and blocking force typically exhibit a trade-off relationship,[40] the SGB-based actuator shows a simultaneous increase in both metrics, indicating improved actuator performance. This enhancement could arise from differences in bending stiffness, as the SGB-based actuator lacks an additional silicone encapsulation layer, which allows for a larger displacement and potentially enables more effective contact with the load cell during force measurements.

Next, we fabricate and characterize SGB-based fluidically-driven soft actuators in various configurations to demonstrate the customizability afforded by SGB. In these designs, a glossy paper sheet of arbitrary geometry is attached to a cylindrical silicone chamber, serving as the constraint layer (Fig. S4, Supplementary Information). The shape of this constraint layer programs the chamber's deformation.[22–24,41] Here, we demonstrate three representative modes: a contraction that restricts longitudinal expansion, a bending that restricts expansion on one side, and a twisting that introduces a helical constraint. Upon pressurization, all actuators achieve their intended motions (Fig. 4d). The contraction type exhibits a 3.2% contraction at 10 kPa, the bending type achieves a 36° tip angle at 8 kPa, and the twisting type generates a 55° twist angle at 8 kPa. For the contraction design, no deformation is observed at pressures below 7 kPa, likely due to the constraint layer dimensions; we expect that deformation at lower pressures can be achieved through geometric redesign. For the bending and twisting types, deformation scales with input pressure within the 0–8 kPa range. These results demonstrate that SGB applies to devices exhibiting diverse deformation modes. Moreover, combining this high degree of design flexibility with the reversible detachment capability suggests that SGB may greatly facilitate rapid redesign, customization, and iterative prototyping of soft robotic devices.

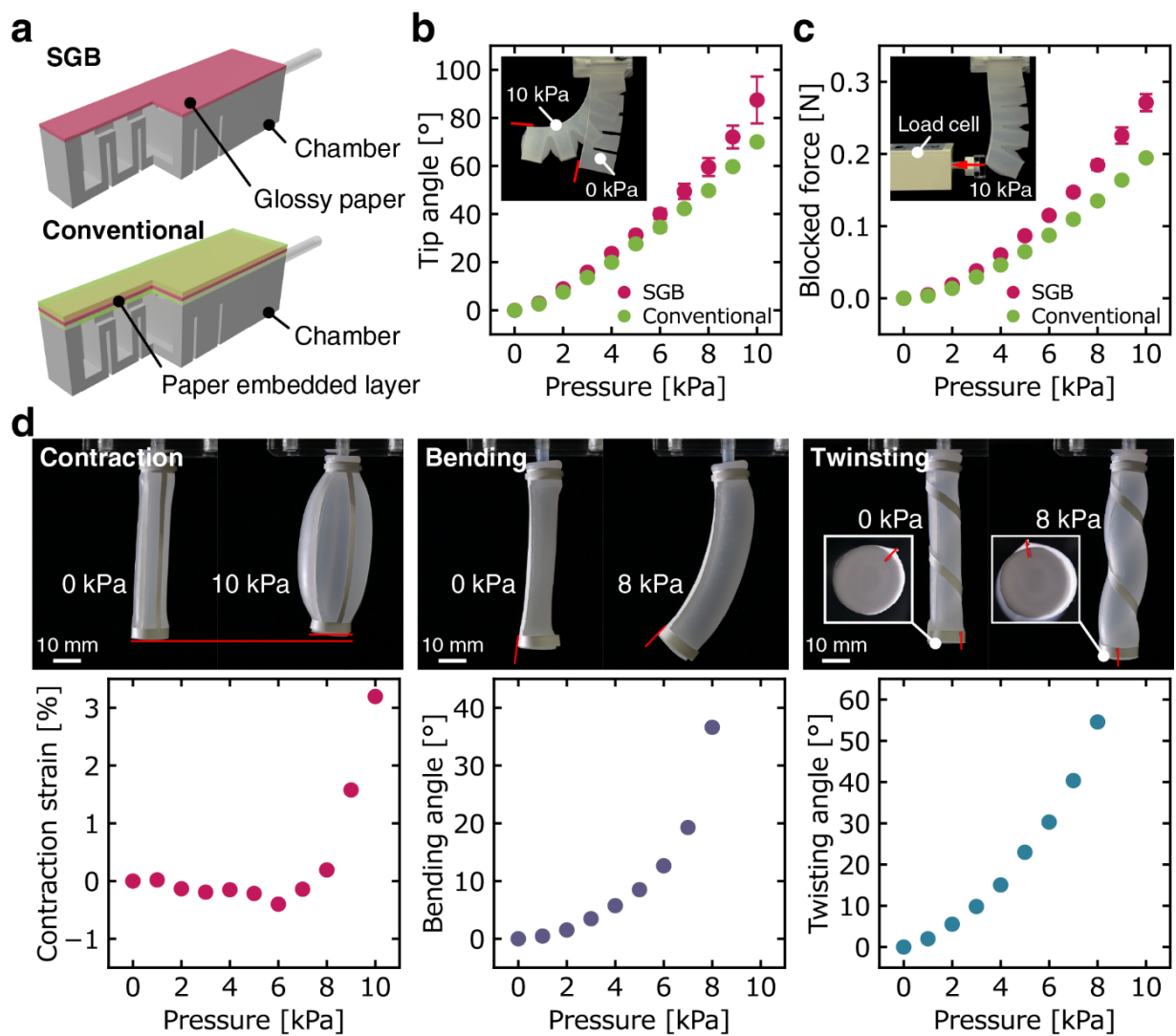

**Fig. 4: Fluidically-driven soft actuators based on SGB. a** Structure of bending actuators fabricated using SGB and a conventional embedded-paper method. **b** Tip angle and **c** blocking force as functions of input pressure. **d** Actuation behavior and corresponding quantitative deformation response of SGB actuators programmed for contraction, bending, and twisting. Values in **b** and **c** represent the mean and the standard deviation ($N = 3$).

**Discussion**

We presented a rapid and reversible adhesive-free bonding strategy that establishes a robust adhesion interface between silicones and glossy papers. Our findings indicate that adhesion strength is governed by the chemical state of the glossy paper coating and the modulus of the silicone, while also offering complete reversibility through solvent immersion. Despite its radically simplified fabrication process, the resulting actuators demonstrate performance metrics that match or surpass those of conventional designs. By modulating the geometry of the glossy paper, we achieved versatile actuation modes, including contraction, bending, and twisting, which underscores the high degree of customizability afforded by this approach. This method is therefore expected to greatly expand the design space for soft devices, promote rapid prototyping, and enable inherent reconfigurability.

**Methods**

*Silicone and glossy paper*: A platinum-catalyzed silicone (Ecoflex 00-30, Smooth-On) was used as the base elastomer. Seven types of glossy papers were employed and labeled as follows: single-sided glossy papers P1 (NME0287-01, Epson), P2 (KL50SCKR, Epson), P3 (QP30A6GSH, Konica Minolta), P4 (GL-101GS50, Canon), P5 (KJ-RG1445N, Kokuyo), and P6 (EJK-GANH50, Elecom); and the front ($P7_a$) and back ($P7_b$) sides of a double-sided glossy paper (KA410PSKD, Epson).

*Fabrication of Specimens*: Specimens were configured based on the "FIN SEAL" format described in ASTM F88. Parts A and B of the silicone were mixed in a 1:1 weight ratio, stirred in a mixer (AR-100, Thinky) at 2000 rpm for 3.5 min, and subsequently defoamed at 2200 rpm for 1 min. The silicone mixture was blade-coated onto a polyethylene terephthalate (PET) film (S10, Toray) using a gap applicator (ZUA2000, Zehntner) and an automatic film applicator (KT-AB4220, Cotec). To achieve a post-curing thickness of 1 mm, the applicator speed and gap were set to 5.0 mm s$^{-1}$ and 1.75 mm, respectively. Curing was performed in an electric oven (DY400, Yamato Scientific) at 80 °C for 20 min. All processes were conducted in a class 10000 cleanroom. After curing, a protective oriented polypropylene (OPP) film (#30, O-cello) was placed on the silicone surface. The silicone film and glossy paper were cut to the specified dimensions: 15 mm × 15 mm for the silicone (using a laser cutter) and 15 mm × 89 mm for the glossy paper (using a disposable scalpel; 2-5726-21, Kai, to prevent laser-induced damage). The protective OPP film was removed, and the silicone film was sandwiched between the glossy sides of two glossy paper sheets. The silicone was aligned with the edges of the glossy papers to create a bonded area of 15 mm × 15 mm and unbonded tabs of 15 mm × 74 mm. Finally, a 100 g weight (approx. 4.4 kPa) was placed on the bonded area for 1 min to establish contact.

*T-peel Tests*: Both ends of the specimen were secured with clamps (SCG-50NA, Shimadzu) and peeled at a speed of 50 mm min$^{-1}$ using a tensile tester (AGS-20NX, Shimadzu). Adhesion strength was calculated by dividing the maximum peel force by the specimen width. The average value of three samples was recorded for each condition. For the releasability investigation, deionized water (compliant with JIS K 0557 A3), 99.5% ethanol (14033-89,

Ikeda Scientific), and tap water (Chofu, Tokyo, Japan) were used. The solutions were filled into high-density polyethylene (HDPE) containers, and the bonded area of the specimens was fully submerged. Thermal stability was evaluated by applying hot air to the peeling interface using a digital heat gun (HG-02Y, Anesty). For cyclic testing, the tensile tester was programmed to pull the specimen to a displacement of 65 mm, followed by 200 reciprocation cycles between 65 mm and 85 mm at a rate of 50 mm min$^{-1}$, during which the adhesion strength was monitored. To investigate the effect of silicone hardness, a softener (Slacker, Smooth-On) was pre-mixed with Part B of a stiffer silicone (Dragon Skin 10 MEDIUM, Smooth-On) before mixing with Part A.

*Analysis of Glossy Paper Surface and Adhesive Interface*: The surface roughness of the glossy papers was measured according to ISO 4287 using a stylus profilometer (Dektak DXT-A, Bruker) equipped with a stylus having a 12.5 μm diameter tip radius. Measurements were taken at three locations in both longitudinal and transverse directions to obtain an average value. Elemental composition was analyzed via X-ray Photoelectron Spectroscopy (XPS; JPS-9200, JEOL). Narrow scan spectra were charge-corrected referencing the C1s peak (284.8 eV), and Shirley background subtraction was applied. For interface observation, a 5 mm × 5 mm sample of silicone bonded to P1 glossy paper was gold-coated (SC-701, Sanyu Denshi) and observed using a SEM (SU5000, Hitachi High-Tech). TEM samples were prepared by trimming the bonded interface (approx. 2 mm × 500 μm), coating with platinum (JFC-1600, JEOL), and thinning at −175 °C using a Focused Ion Beam (FIB; JEM-9310FIB, JEOL) and a Helios 650 (FEI) system for surface protection. STEM images, EELS mapping, and EELS profiles were acquired using a TEM (JEM-ARM200F, JEOL). Backgrounds for EELS profiles were determined by extrapolating the pre-edge spectrum and subtracting it from the signal.

*Fabrication of Actuators*: Bending actuator chambers were fabricated by casting the silicone into molds 3D-printed from polylactic acid (PLA) filament (Bambu Lab) using an FDM printer (P1S, Bambu Lab). The conventional strain-limiting layer was formed by casting silicone onto a 1-mm-thick laser-cut polymethyl methacrylate (PMMA) plate. A glossy paper layer was embedded into the strain-limiting layer, which was then bonded to the chamber layer while the silicone remained in a semi-cured state. Balloon actuator chambers were fabricated similarly using a PLA outer mold and a UV resin inner mold (Form4 Clear Resin V5, Formlabs) printed via stereolithography (SLA) (Form4, Formlabs). The inner mold was polished with sandpaper (#400–#9800). All silicones were mixed using a mixer, cast, and then cured in an electric oven at 50 °C for 60 min. Detailed dimensions and fabrication diagrams are provided in Fig. S3, S4 (Supplementary Information).

*Characterization of Actuators*: Actuators were driven by compressed air (ACP-25SLAA, Takagi) controlled by an electro-pneumatic regulator (ITV1050-31F2L, SMC). Displacement was recorded with a camera (Z5, Nikon) equipped with an AF-S Micro NIKKOR 60mm f/2.8G ED lens (Nikon) and quantified using ImageJ software. The blocking force at the tip of the bending actuator was measured using a load cell (DPU-2N, Imada) connected to a digital multimeter (2100, Keithley).

*Use of AI Tools:* ChatGPT (version 5.2, OpenAI) was used solely for language editing support during manuscript preparation. All scientific content, data interpretation, and original writing were independently developed by the authors. All AI-assisted outputs were thoroughly reviewed and verified by the authors.

**Data Availability**

All data needed to evaluate the conclusions presented in this paper are included in the article and/or Supplementary Information.

**Acknowledgements**

This work was supported by the JSPS KAKENHI Grant-in-Aid for Scientific Research (Grant Numbers 23K26072 and 23H01377), the JSPS Research Fellowship for Young Scientists (DC2), the JST Fusion-Oriented Research for Disruptive Science and Technology program (Grant Number JPMJFR2126), JST SPRING (Grant Number JPMJSP2131), and Advanced Research Infrastructure for Materials and Nanotechnology in Japan (ARIM), MEXT (Grant Number JPMXP1224NM0224).

**Author Contributions**

TS, KM, and JS conceived of the main idea. TS, KM, AS, and JS designed the experiments. TS and KM conducted the experiments and analyzed the data. TS generated and edited the figures and media files. TS wrote the manuscript. AS and JS supervised the project. TS, KM, AS, and JS edited the manuscript. All authors reviewed the manuscript.

**Competing Interests**

The authors declare that they have no competing interests.

**Ethics, Consent to Participate, and Consent to Publish declarations**

Not applicable.

# Supplementary Information

**Instant and Reversible Adhesive-free Bonding Between Silicones and Glossy Papers for Soft Robotics**

Takumi Shibuya, Kazuya Murakami, Akitsu Shigetou, and Jun Shintake*

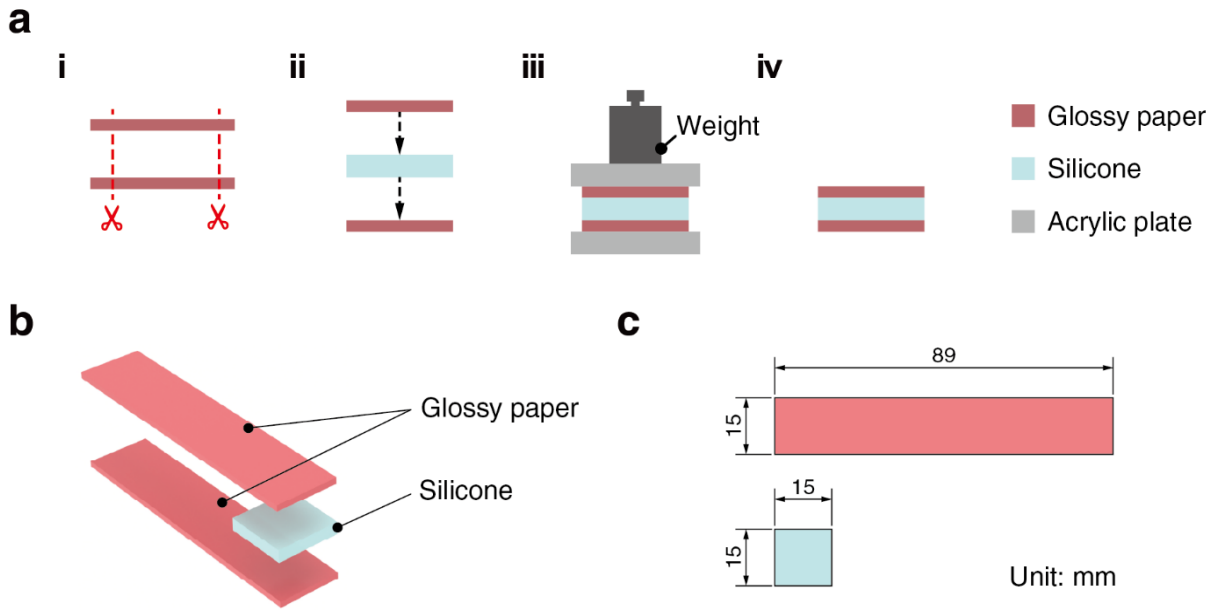

**Fig. S1:** Specimens for evaluating adhesion characteristics. **a** Specimen fabrication process. **b** Specimen configuration. **c** Specimen dimensions.

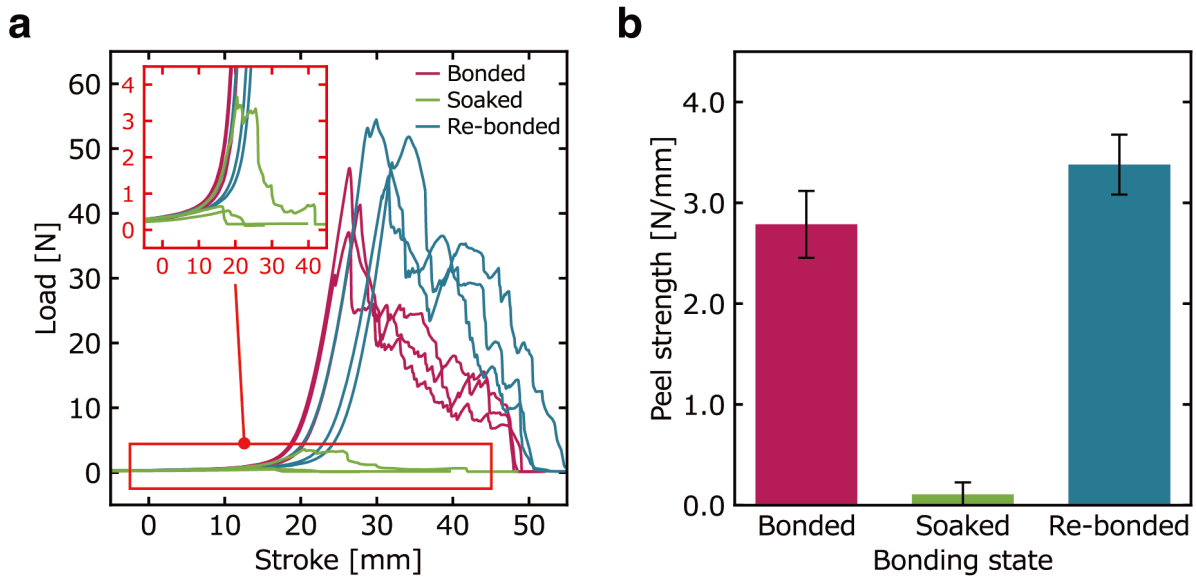

**Fig. S2:** Adhesion strength for different bonding states obtained from three samples each. **a** Force-stroke response. **b** Peel strength calculated from maximum load. Values represent the mean and the standard deviation ($N = 3$).

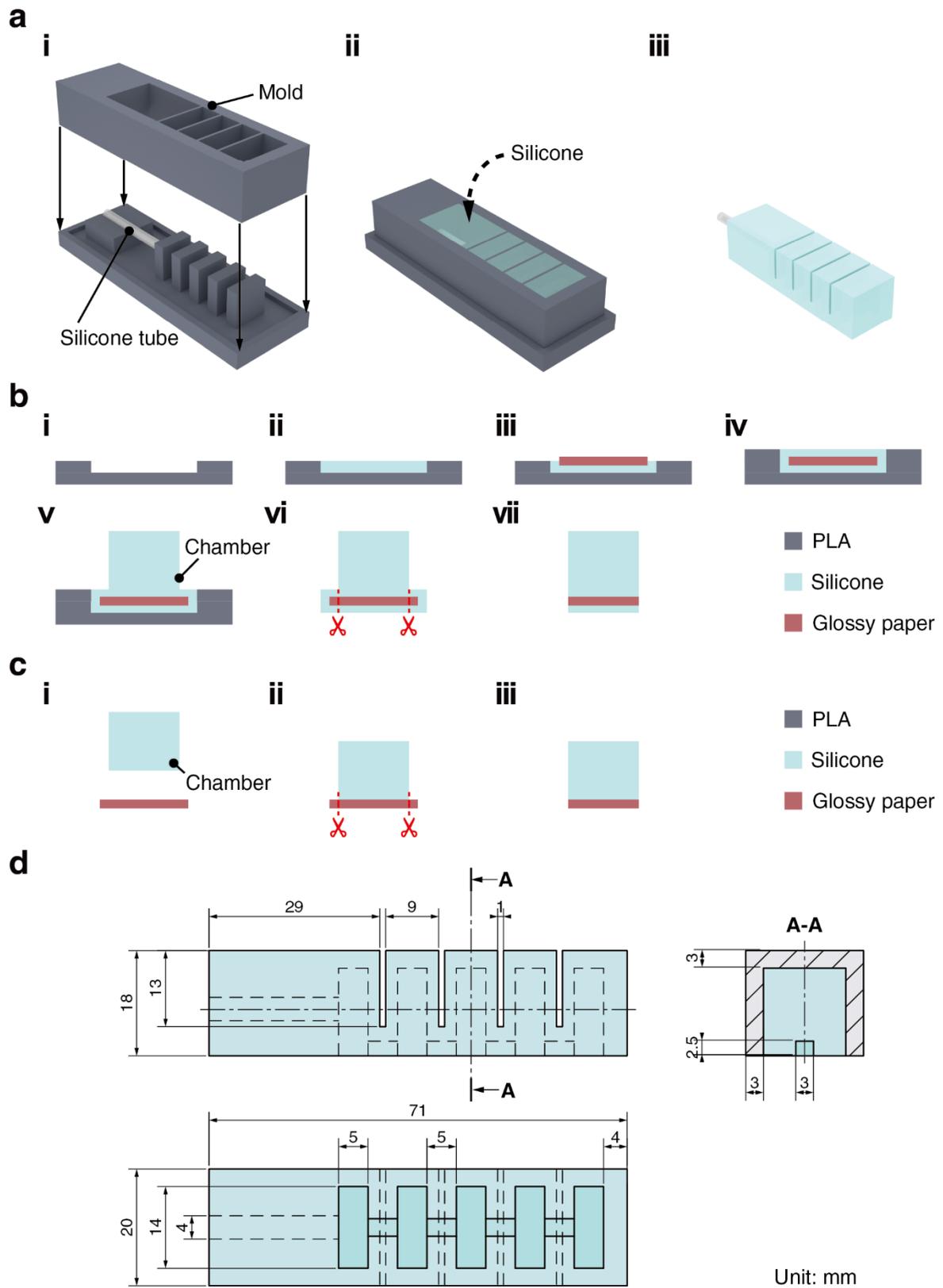

**Fig. S3:** Description of the bending soft actuator. **a** Fabrication of the actuator chamber. **b** Bonding of the rigid layer via the conventional process. **c** Bonding via the proposed method. **d** Dimensions of the silicone chamber.

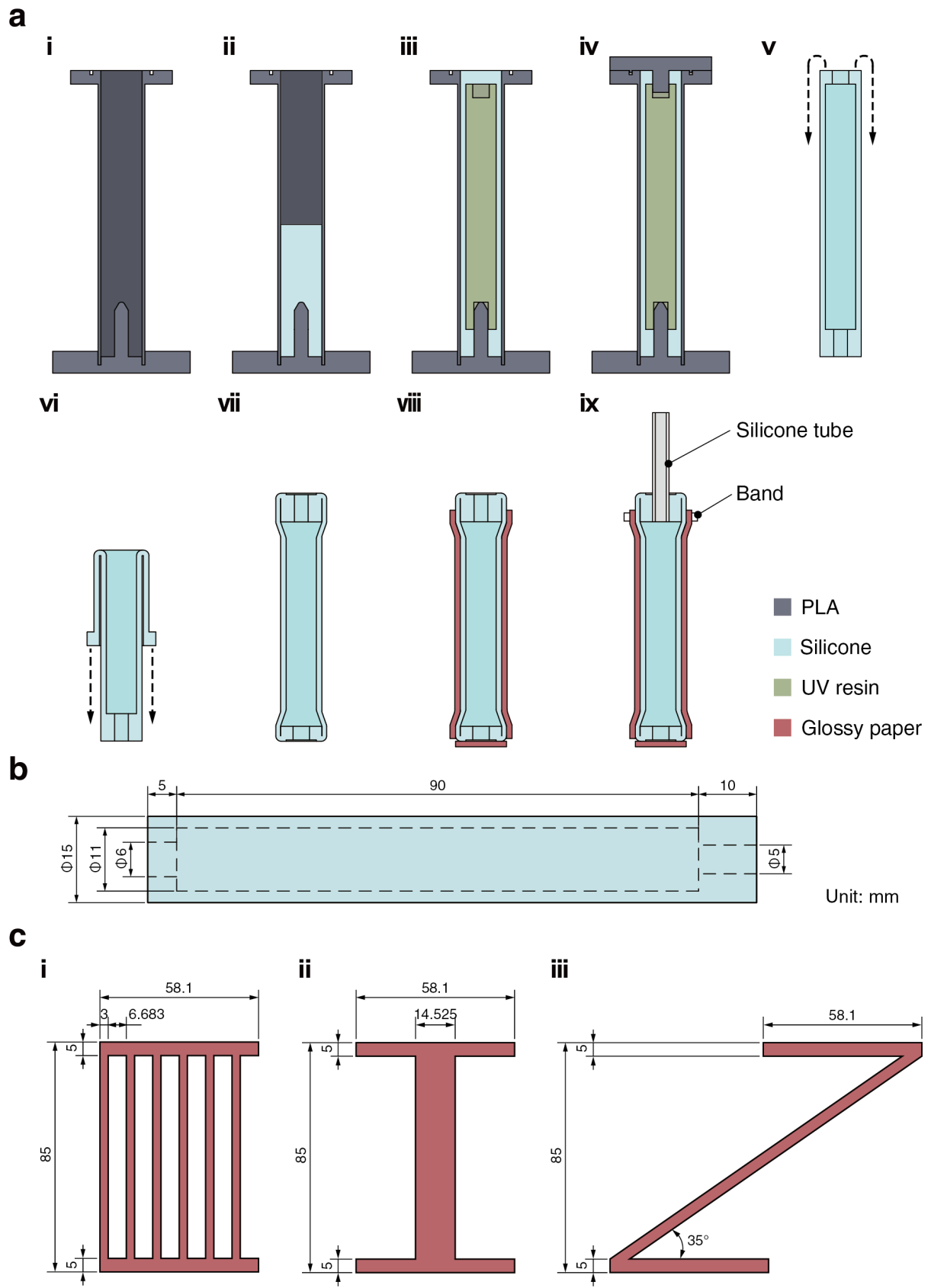

**Fig. S4:** Description of the programmed soft actuator. **a** Fabrication of the actuator. **b** Dimensions of the cylindrical silicone chamber and **c** constraint layer.

**Video S1:** Instant and robust locking. Adhesion phenomenon observed upon contact between a silicone piece and glossy paper. The adhesion is maintained even when tension is applied to the silicone.

**Video S2:** Dynamic loading. A 500 g weight attached to the glossy paper via a clip. The bonding remains stable even under dynamic loading.

**Video S3:** Human body weight loading demonstration. A silicone–glossy paper composite loop with a bonding area of 89 mm × 90 mm supports a human weight of ~50 kg without delamination.

**Video S4:** Rapid fabrication of soft actuator. A pneumatic bending actuator is assembled simply by contacting silicone chambers with a glossy paper strain layer. The resulting silicone–glossy paper interface creates sealed chambers capable of sustaining the pneumatic pressure required for bending actuation.